\documentclass{llncs}

\usepackage{amsmath} 
\usepackage{xcolor}
\usepackage{booktabs}
\usepackage{url}
\usepackage{hyperref}
\usepackage{graphicx}
\usepackage{subfigure}

\title{Patch Selection for Melanoma Classification}

\author{Guillaume Lachaud \inst{1,2} \and Patricia Conde-Cespedes
\inst{1,3}             
\and Maria Trocan \inst{1, 4}}
\institute{ISEP - Institut Supérieur d’Électronique de Paris.\\
10 rue de Vanves, Issy les Moulineaux, 92130-France\\
\and \email{glachaud@isep.fr} \and \email{pconde@isep.fr}
\and \email{maria.trocan@isep.fr} }

\begin{document}

\maketitle

\abstract{In medical image processing, the most important information is often
located on small parts of the image. Patch-based approaches aim at using only
the most relevant parts of the image. Finding ways to automatically select the
patches is a challenge. In this paper, we investigate two criteria to choose
patches: entropy and a spectral similarity criterion. We perform experiments at
different levels of patch size. We train a Convolutional Neural Network on
the subsets of patches and analyze the training time. We find that, in addition to requiring less preprocessing time, the classifiers trained on the datasets of patches selected based on entropy converge faster than on those selected based on the spectral similarity criterion and, furthermore, lead to higher accuracy. Moreover, patches of high entropy lead to faster convergence and better accuracy than patches of low entropy.}


\section{Introduction}
With the development of better machine learning methods driven by deep learning,
there have been many successful applications of neural networks in medical image
processing, such as biomedical image segmentation or cancer diagnosis
\cite{DBLP:journals/corr/abs-1709-02250}. This is the case in cancer diagnosis
and prognosis, with applications in breast cancer
\cite{yalaDeepLearningMammographybased2019}, lung cancer
\cite{marentakisLungCancerHistology2021a} and skin cancer
\cite{estevaDermatologistlevelClassificationSkin2017}.

Medical images can be widely different depending on their source, such as CT
(Computed Tomography) scans, MRI (Magnetic Resonance Imaging) images,
dermoscopy, etc. While classification is usually performed on the whole images,
medical images can have extremely high resolution, e.g. gigapixels for skin
tissue images, which makes it more time efficient to train on subsets or patches
of images. Additionally, it can enhance a classifier performance
in some settings. For example, in \cite{7780635}, the authors argue that cancer
subtypes are distinguished at the image patch scale. Patch-based classification
is also used in \cite{ROY201990} for breast histology. More applications of
patch-based applications are introduced in \cite{5771116} and \cite{6690248}.

A judicious choice of patches reduces the importance of noise and focuses on the most
important parts of the image.  Two approaches of selecting the patches used for
classification are to score the patches individually based on a given metric, or
to compare each patch with the other patches of an image and rank the similarity
between the patches. In the first approach, the patches can be scored using
entropy, while the second approach relies on a similarity measure between
images.

On one hand, entropy is used in information theory as a way to quantify the
level of information of an object. Higher entropy means that there is more
information in the object. For instance, a random noise image has high entropy
while a unicolored one has very low entropy. Entropy plays an important role in
data compression where it provides the lower bound on the storage required to
compress an object without loss of information
\cite{DBLP:journals/bstj/Shannon48}. Entropy can also be used for object
reconstruction using the principle of maximum entropy, which aims at selecting
the most uniform probability distribution amongst multiple candidate
distributions. It can be used for image reconstruction where the candidates are
the set of missing pixels \cite{skilling1984maximum}. It applies to text data as
well \cite{nigam1999using}. Entropy can also be used in image texture analysis
\cite{DBLP:journals/neco/ZhuWM97} and texture synthesis. Selecting patches using
entropy was explored in \cite{DBLP:conf/iccci/LachaudCT21}.

On the other hand, the Mean Exhaustive Minimum Distance (MEMD) is a criterion
that was introduced in \cite{havlicekTextureSpectralSimilarity2019} to compare
two images by trying to find the best pairing of pixels from the first and the
second image; the criterion score then indicates how similar the images are. A
low score indicates that the images are similar, and a high score that the image
are different. This can be extended to comparison between a patch and several
patches by averaging the scores. 

In this paper we study the training time and the accuracy of these two criteria for patch-based
binary classification. The data we use comes from the ISIC (International Skin
Imaging Collaboration) archive.  \footnote{The data is publicly available at
\url{https://www.isic-archive.com}} This consortium was created to improve the
fight against skin melanoma cancer by improving computer-aided diagnosis. The
consortium has held an annual challenge since 2016
\cite{gutmanSkinLesionAnalysis2016}. Starting from 2019, the challenges are
centered around dermoscopic image multi-class classification. The best team on
the 2019 challenge \cite{gessertSkinLesionClassification2020a}, investigated
patch-based classification on the HAM10000 dataset
\cite{tschandlHAM10000DatasetLarge2018}, where information from several patches
is combined via an attention-based mechanism.

The paper is divided as follows. In Section~2 we present the dataset and the
preprocessing we perform on the data. We also introduce the entropy and MEMD
criterion that we use in our experiments. In Section~3 we present the results of
our experiments and we conclude the paper in Section~4.


\section{Materials and Methods}

In this section we describe the dataset, the criteria of entropy and
Mean-Exhaustive Minimum Distance (MEMD) we use, and the network architecture
that is trained on the data.

\subsection{Dataset description and preprocessing}%
\label{sub:dataset_description_and_pre_processing}

The ISIC archive database comprises skin lesion images associated with a label
indicating the status of the lesion. The image
resolution is arbitrary. The archive provides an API to retrieve the images and
their metadata, as well as the mask of the region of interest when an expert has
created one. The total number of patches created is presented in
Table~\ref{tab1}. We perform binary classification on patches of the
images. Our target variable is a categorical variable with two possible values:
\textit{benign}, or \textit{malignant}.

The preprocessing steps are:

\begin{enumerate}
    \item We download images from the ISIC archive, as well as the masks that
    are annotations from experts and indicate the lesion location.
\item All the malignant images with a mask are selected. The same number of benign images is
sampled out of all the benign images.
\item The region of interest is divided in square patches of width 32, 64, 128 and 256. The region
of interest is defined by the downloaded masks.
\item The entropy of each patch is computed, and we use these values to extract a subset of patches.
This is explained in section \ref{sub:entropy}.
\item For each image, we compute a spectral measure of similarity between a patch and all the
    other patches of the image; we use this measure to extract a subset of
    patches. The details are in section
    \ref{sub:mean_exhaustive_minimum_distance_criterion}.
\item Finally, a classifier is trained on all the datasets we have created in the
    two previous steps.
\end{enumerate}

\begin{table}[h]
    \begin{center}
\small
\caption{Number of patches for each patch size\label{tab1}}
\begin{tabular}{cc}
\toprule
\textbf{Patch size}	& \textbf{Number of patches}\\
\midrule
$32\times 32$		& $4,889,969$			\\
$64\times 64$		& $1,173,052$			\\
$128\times 128$		& $270,821$			\\
$256\times 256$		& $58,253$			\\
\bottomrule
\end{tabular}
    \end{center}
\end{table}

We divide the images in three groups: $90\%$ of the images are in the train set,
with $20\%$ of the train set reserved for validation; the remaining $10\%$
constitutes the test set.

\subsection{Entropy}%
\label{sub:entropy}

We use the Shannon entropy \cite{DBLP:journals/bstj/Shannon48}. It is defined by
the formula

\begin{equation}
    H = \sum_{k=1}^M p_k \log(p_k)
\end{equation}

where we sum across all the pixel intensities, i.e. from 0 to $M=255$. $p_k$ is
the probability a pixel in the image is at intensity $k$. The entropy ranges
from $0$ to $\log_2(255)\approx 8$. Because there is no consensus on how to
compute the entropy for multi channel images, we convert our RGB images to
grayscale. The conversion process is defined in the ITU-R Recommendation
BT.601-2.

Using the entropy, we extract two datasets for each patch size:
\begin{itemize}
\item 	a \textit{low} dataset, whose patches are all the patches that rank below
    the 15-th and the 30-th quantile of entropy with respect to the other patches of the same image.
\item a \textit{high} dataset, with entropy above the 85-th \and{and the 70-th} quantile.
\end{itemize}

\subsection{Mean Exhaustive Minimum Distance (MEMD) criterion}%
\label{sub:mean_exhaustive_minimum_distance_criterion}

The first methods of similarity measure usually consisted in computing certain
features on a given image, such as the Haralick features
\cite{haralickTexturalFeaturesImage1973}, and then comparing the features
obtained for different images. More recent techniques dealing with the
structural similarity in textures have been proposed in
\cite{zujovicStructuralSimilarityMetrics2009} and
\cite{qinSimilarityMeasureLearning2004}. Handling color or hyperspectral images
is often done using histograms
\cite{yuanFactorizationBasedTextureSegmentation2015}, but histograms require a
large amount of data to get good estimates of the spectral distribution. A new
criterion to evaluate the similarity of two images was proposed in
\cite{havlicekTextureSpectralSimilarity2019}. This approach does not require
histograms and generalizes to any number of channels.

Following the notation from \cite{havlicekOptimizedTextureSpectral2021}, let $A$
and $B$ be two images, which can have multiple channels. Let $M=\min(\#A,\#B)$,
with $\#A$ and $\#B$ the number of pixels in $A$ and $B$. Let $\langle A
\rangle$ be the set of pairs of coordinates for the pixels of A, and $U$ the
unprocessed pairs of coordinates of pixels of B. Let $\rho$ be the distance
induced by a vector metric. $A_{i,j}$ denotes the pixel of $A$ at coordinates
$(i,j)$; the channels dimension is implied. Similarly, $B_{k,l}$ is the pixel of
$B$ at coordinates $(k,l)$.  The MEMD criterion $\zeta$ is defined by
Equation~2.

\begin{equation}
    \zeta(A,B) = \frac{1}{M} \sum_{(i,j)\in \langle A \rangle} \min_{(k,l) \in
    U} \{ \rho(A_{i,j}, B_{k, l}) \}
\end{equation}

The lower the score is, the more similar images $A$ and $B$ are. Inversely, the
higher the score, the higher the difference between the two images. The score
can take values between 0 and 255. A score of 0 happens when we compare one
image to itself; a score of 255 happens when we compare a white image with a
black one.

To improve the computation time, \cite{havlicekOptimizedTextureSpectral2021}
suggested that the pixels of both the images be sorted with respect to the
chosen norm. Finding the minimum distance between the pixels of the two images then
comes down to choosing the closest unprocessed neighbour in the sorted array. In
the special case where $A$ and $B$ are of the same size, we can simply match the
first element of the sorted pixels of $A$ with the first of element of the
sorted pixels of $B$, and so on.

\begin{figure}[h]
    \begin{center}
        \subfigure[$32 \times 32$ patches]{\includegraphics[width=0.45\textwidth]{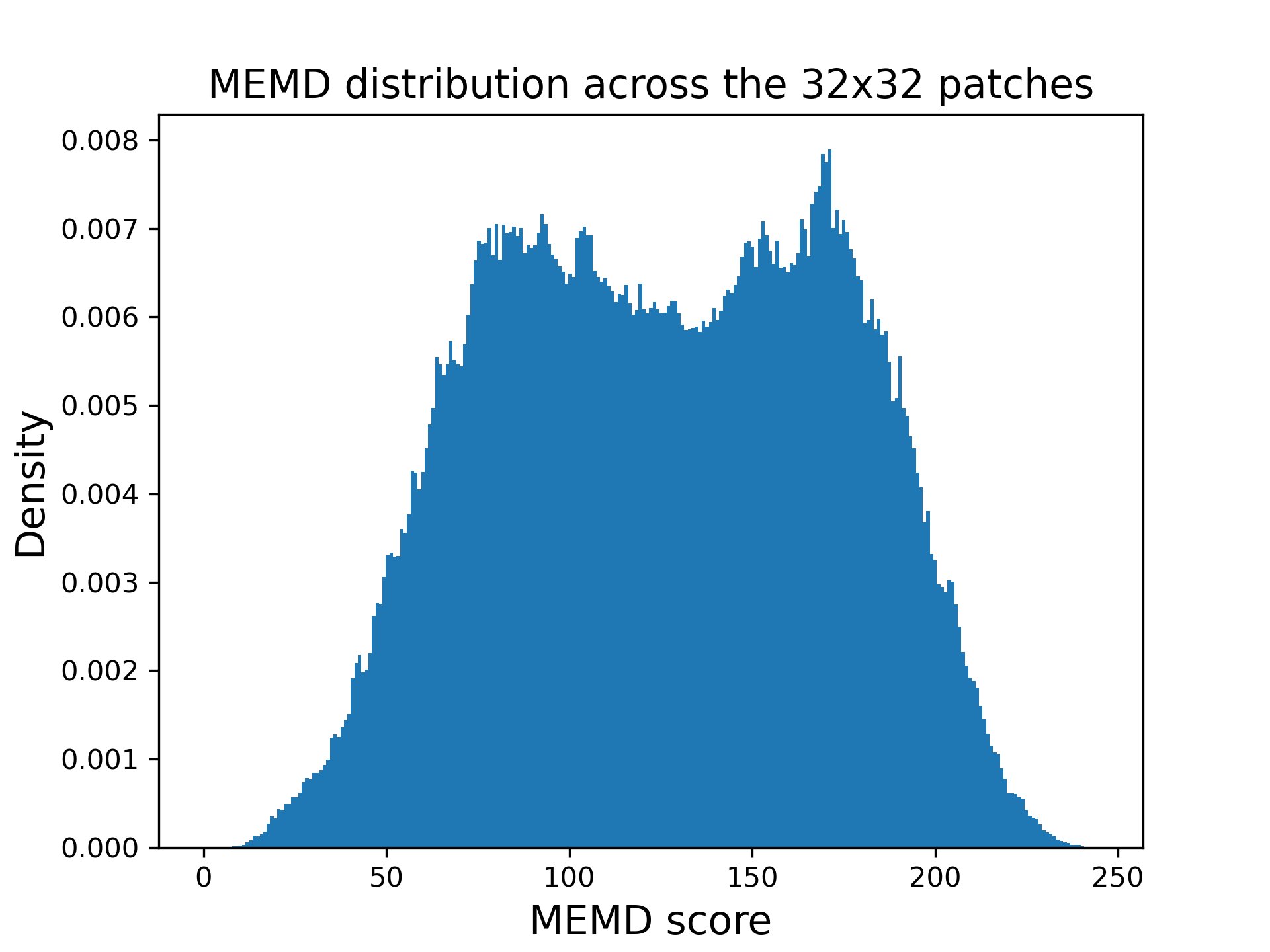}} 
        \subfigure[$64 \times 64$ patches]{\includegraphics[width=0.45\textwidth]{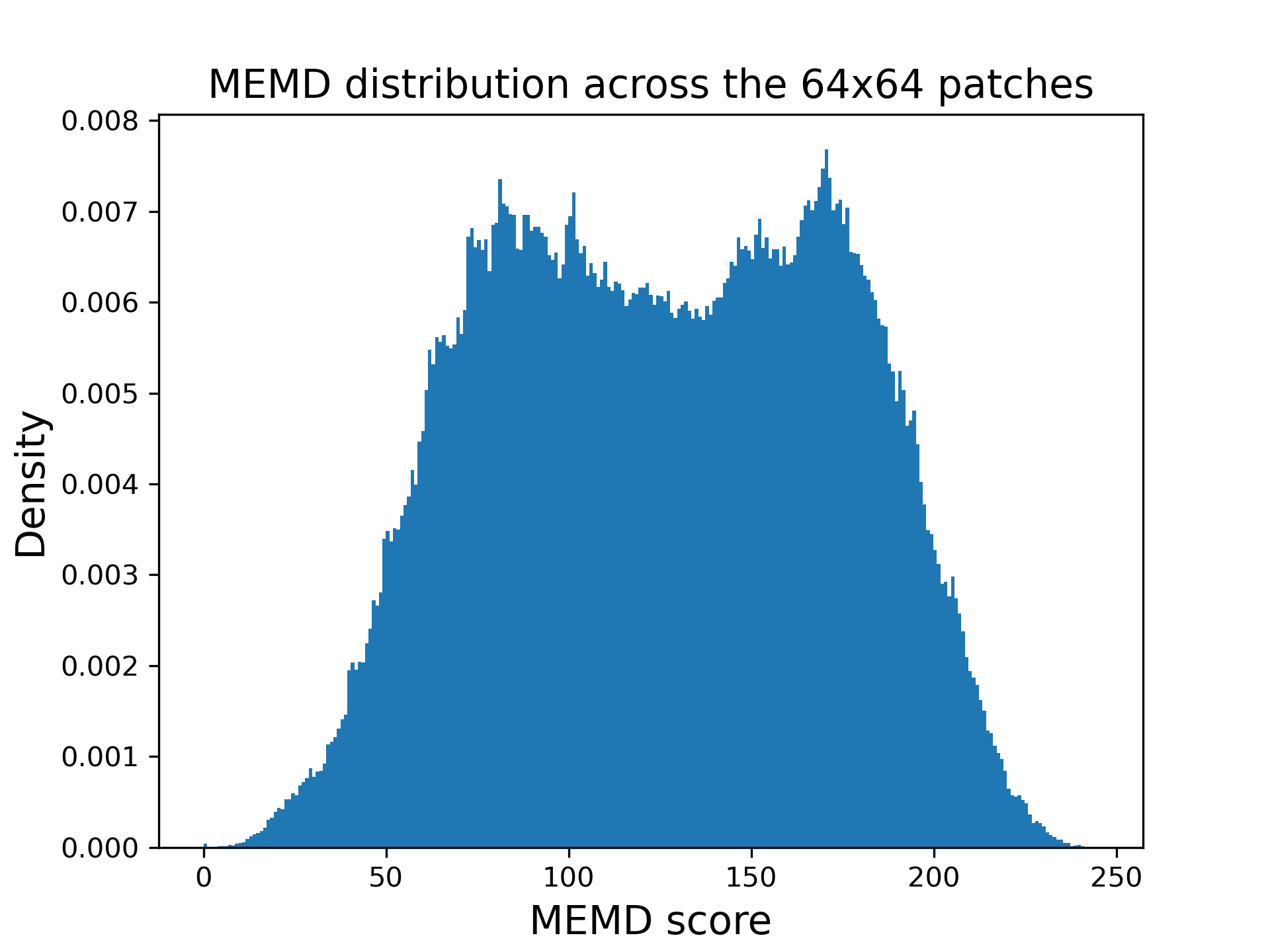}} 
        \subfigure[$128\times 128$ patches]{\includegraphics[width=0.45\textwidth]{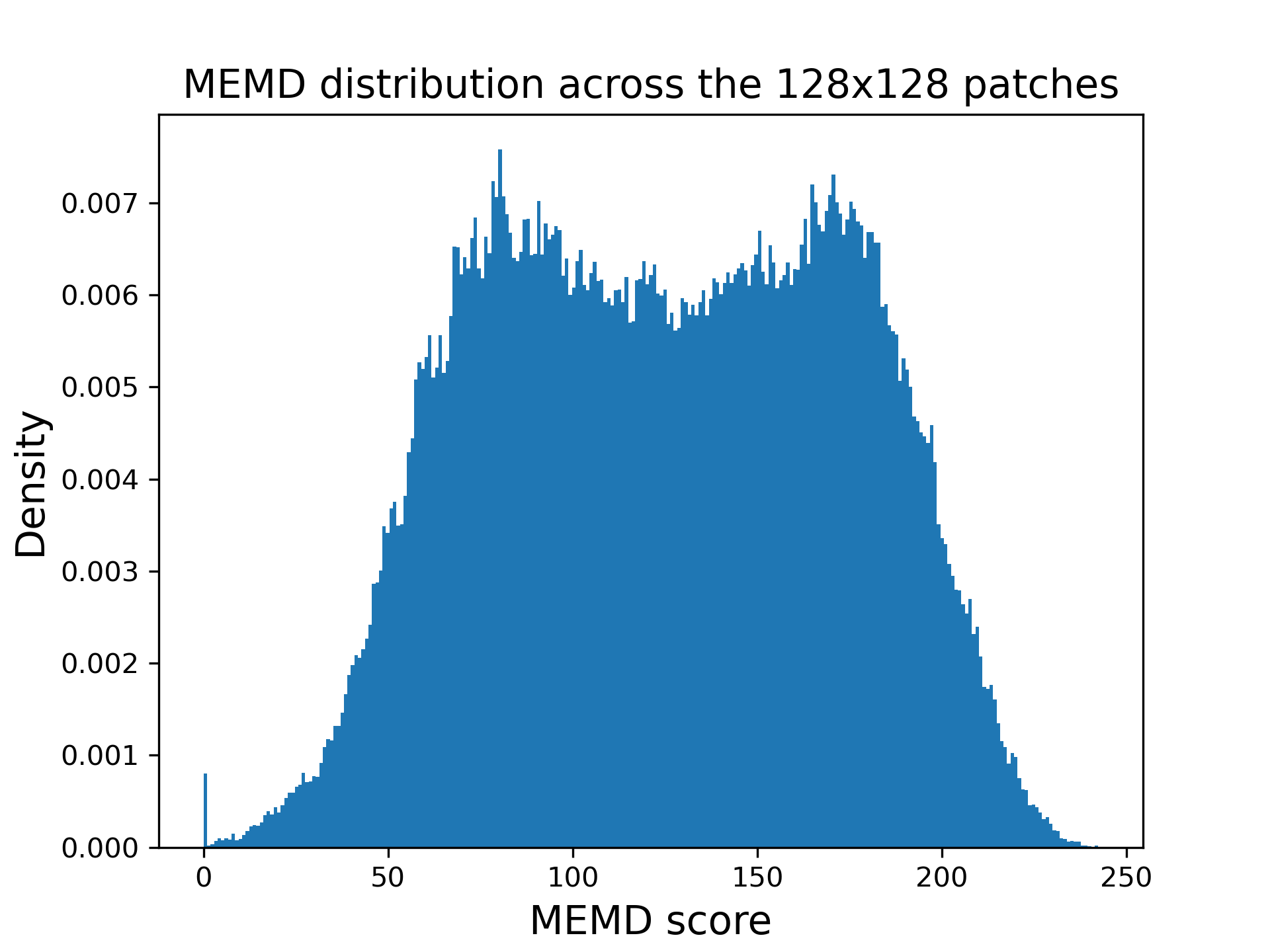}}
        \subfigure[$256 \times 256$ patches]{\includegraphics[width=0.45\textwidth]{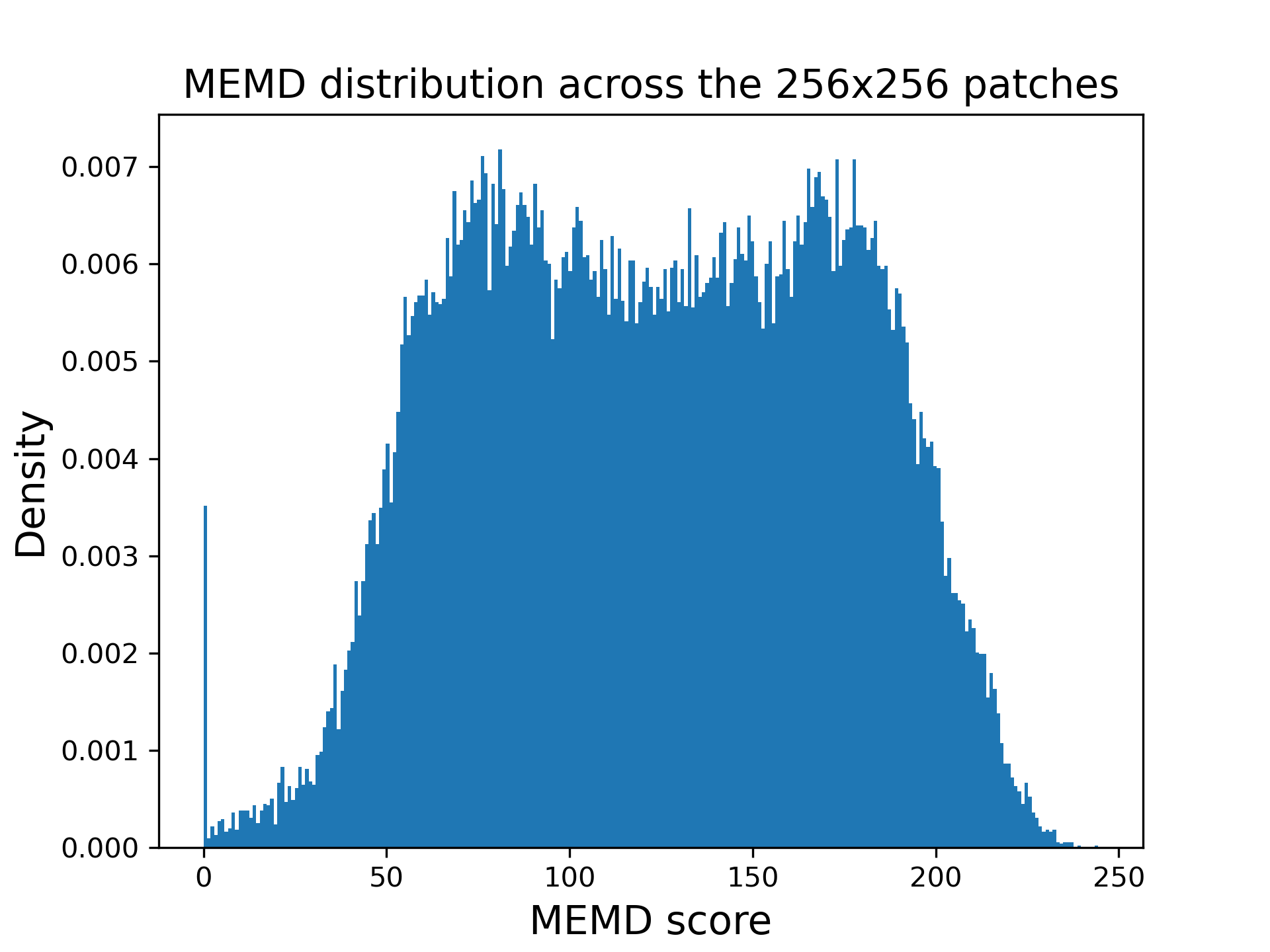}}
\caption{Distribution of MEMD score for different patch sizes\label{fig:memd}}
        \label{fig:entr}
    \end{center}
\end{figure}

We compute the MEMD score of each patch with respect to all the other patches of
the same image, and we average the scores. Figure~\ref{fig:memd} shows the
distribution of the MEMD score at varying patch sizes. We observe two peaks. The
peak on the left corresponds to the patches that are representative of the
overall image, and the peak on the right corresponds to the patches that are
more unique. The reason why we only have two peaks is that the images of the
lesion all share similar elements: a little bit of skin, the lesion, and some
noise such as hair, a ruler, etc. The distinction between the lesion and the
skin is quite drastic, meaning that few patches are going to be equally similar
to skin and lesion. The variation in scores is in part due to the different
number of patches per image. The more patches an image has, the less extreme the
MEMD score of the patches will be. The patches with a score of 0 are from images
that have only one patch. This happens for big patch sizes where the region of
interest is too small to get more patches.

Similarly to what was done in Section\ref{sub:entropy}, we create datasets using
the same quantiles for the MEMD score.

In the rest of the paper, we use the max norm for $\rho$, i.e. $\rho : x \mapsto
\left\lVert x \right\rVert_{\infty} = \max_{i} \lvert x_i \rvert$, where $x_i$
are the coordinates of $x$. The distance induced by the max norm is $(x, y)
\mapsto \left\lVert x-y \right\rVert_{\infty}$. Because the sorting of the
pixels is done based on the norm of a single pixel, and the min is computed
using the distance between two pixels, the optimization via sorting is not
compatible with pixels with multiple channels. Indeed, let $p_1 = [135, 18,
89]$, $p_2 = [130, 16, 86]$ and $p_3 = [12, 134, 1]$. If we sort the pixels by
the max norm, we get $P=[p_2, p_3, p_1]$.  Selecting the closest matching pixel
using the proposed method in \cite{havlicekOptimizedTextureSpectral2021} would
make us pair $p_2$ with $p_3$, which leads to $\zeta(p_2, p_3) = 118$. But $p_2$
and $p_1$ are clearly a better match, with $\zeta(p_2, p_1) = 5$. To alleviate
the complications imposed by the multiple channels, we convert the images to
grayscale before computing the MEMD score. Since the grayscale image has only
one channel, the optimization via sorting works.

The computation of the average MEMD of all the patches of an image has $O(m^2)$
with respect to $m$, the number of patches in the image. There is a trade-off
between space and time complexity, where vectorizing part of the process using
higher order tensors allows for faster computation but requires more space.

\subsection{Network Architecture}%
\label{sub:network_architecture}

For the choice of classifier, we follow \cite{yilmazBenignMalignantSkin2020} and
\cite{favoleMelanomaDetectionUsing2020} who found that ResNet50 achieved the
best results for the same task and dataset. ResNet50
\cite{heDeepResidualLearning2016} is a 50-layer convolutional neural network (CNN)
that was proposed to alleviate the problem of vanishing and exploding gradients
\cite{DBLP:journals/jmlr/GlorotB10} by introducing the notion of
\textit{residual units}.

With enough computing resources, ResNets can have as many layers as we want,
e.g. 101 or 152 layers. We use the 50-layer version, which we adapt to binary
classification by removing the last layer and replacing it with a max pooling
layer followed by a Dense layer and a \textit{sigmoid} activation. 

We use the Adam optimizer \cite{Kingma2015AdamAM}. We set the learning rate to
$0.001$ and we use a \textit{binary cross-entropy loss} for training.

We train the model for 10 epochs, each epoch representing a full pass through
the train set.To mitigate overfitting, the training stops if the validation loss
does not decrease after 3 consecutive epochs.

Additionally, we investigate combining predictions from several patches of an
image to classify the image. We train a Resnet for 10 epochs and choose the
weights that result in the best validation loss. To classify an image from the
test set, we individually classify its patches and aggregate the results. Let
$\mathcal{P}_i$ be the set of patches from an image $I_i$, $|\mathcal{P}_i|$ the
number of patches selected from the image, $f$ be the classifier that maps a
patch to $0$ for a benign patch and $1$ for a malignant one. The prediction
$\hat{y}$ is given by the Equation~3.
\begin{equation}
    \hat{y}_i = \begin{cases}
	0 & \text{if }
	\left(\frac{1}{|\mathcal{P}|}\sum_{p\in\mathcal{P}_i}f(p)\right) < 0.5 \\
	1 & \text{otherwise}
    \end{cases}
\end{equation}

\section{Experimental Results}

The experiments were performed with an Nvidia Titan XP GPU. The code is written
in Python and Tensorflow. The Pillow library was used for computing the entropy.

To make our results robust against the random initialization of the model
parameters, we train 10 instances of a ResNet50 per dataset. Results of the
experiments with the entropy datasets are presented in Table~\ref{tab3}, and
those performed on the MEMD datasets are presented in Table~\ref{tab4}. The low entropy dataset contains the $15\%$ patches with the lowest entropy, and the high entropy dataset contains the
$15\%$ patches with the highest entropy. Correspondingly, the low MEMD dataset contains the $15\%$ patches with the lowest MEMD score, and the high MEMD dataset contains the $15\%$ patches
with the highest MEMD score.
Additional results for datasets with intermediate entropy are presented in
\cite{DBLP:conf/iccci/LachaudCT21}.

\begin{table}[h]
    \begin{center}
\small
\caption{Quantiles of training time for datasets of different entropy and patch
size\label{tab3}}
\begin{tabular}{c c c c c}
\toprule
& &  \multicolumn{3}{c}{Quantile of training time (in seconds)} \\
patch size & \quad entropy &  \quad 30 \quad & \quad 50 (median) & 70 \\
\midrule
32 & high & 1350.7 & 2013.2 & 2781.4 \\
32 & low &  1534.9 & 2906.7 & 3078.5 \\
\midrule
64 & high & 291.0 & 382.9 & 441.9 \\
64 & low & 290.6 & 338.3 & 414.2 \\
\midrule
128 & high & 155.0 & 204.6 & 220.0 \\
128 & low & 204.8 & 255.0 & 255.4 \\
\midrule
256 & high & 142.4 & 152.2 & 189.7 \\
256 & low & 189.6 & 226.4 & 226.5 \\
\bottomrule
\end{tabular}
    \end{center}
\end{table}

\begin{table}[h]
\small
\begin{center}
\caption{Quantiles of training time for datasets of varying MEMD score and patch
size\label{tab4}}
    \begin{tabular}{c c c c c}
\toprule
    &     &  \multicolumn{3}{c}{Quantile of training time (in seconds) }\\
	patch\_size & memd\_score & 30  &  50 (median) &  70  \\
\midrule
32  & high &        3150.4 &  3254.8 &         3258.9 \\
    & low &        3256.4 &  3260.0 &         3260.9 \\
\midrule
64  & high &         465.3 &   495.1 &          527.0 \\
    & low &         564.4 &   691.9 &          986.3 \\
\midrule
128 & high &         241.5 &   281.9 &          387.9 \\
    & low &         256.6 &   357.7 &          373.1 \\
\midrule
256 & high &         189.7 &   245.2 &          264.4 \\
    & low &         215.4 &   226.7 &          275.2 \\
\bottomrule
\end{tabular}
\end{center}
\end{table}

Regarding the entropy datasets, we observe a tendency of faster convergence for
datasets with higher entropy compared to datasets with lower entropy.
Lower entropy means that the distribution of pixel intensity concentrates on
fewer pixels than it does for higher entropy. This concentration makes for
smoother textures, which might be harder for the classifier to learn. Higher
entropy datasets have more salient features that more discernible and thus more
easily learnable by the network.

As for the MEMD datasets, the dataset composed of patches with higher score
tends to converge faster than the dataset with lower score. This might be
explainable by the fact that a low MEMD score means a high similarity of the
patch with the rest of the image, while a high score indicates a distinctive
spectral texture compared with the other patches of the same image. Thus, the
higher score patches capture the more unique features of the lesion, while the
lower score patches are more representative of the overall texture of the
lesion. The high representativeness of a patch might extend to patches of low
score from another image, while the unique features are probably different
between images. Therefore, the dataset with high score is richer in more unique
patches, which provide more information than the similar patches contained in
the lower score dataset. This, in turn, makes the network training converge
faster for the dataset with higher score patches.

These interpretations are borne out by the results of the experiments presented
in Table~\ref{tab5}. For the $128\times 128$ patches, the accuracy does not improve when
we select more patches: it stagnates around $50\%$. This indicates that this
patch size is too small to properly discriminate the lesions. The problem is not
about the number of patches but about the fact that small patches do not contain
enough information to determine the status of the lesion. We believe that this
situation holds also for even smaller patches, e.g. $32\times 32$ or
$64\times64$ patches. Conversely, for the case of $256\times 256$ patches, we
remark that using too few patches results in very low accuracy (around $25\%$);
however, the accuracy increases considerably  when we select more patches
($30\%$ of $15\%$), achieving $71\%$ accuracy for patches of high entropy.
This accuracy is similar to the $74\%$ accuracy obtained by the
authors of \cite{favoleMelanomaDetectionUsing2020} when training on the whole
region of interest with a ResNet50.

\begin{table}[h]
    \begin{center}
\small
\caption{Test accuracy (in percentage) for the different datasets. For a
given patch size, the test images are the same for each method.\protect\label{tab5}}
    \begin{tabular}{c c c c c}
\toprule
Dataset & low MEMD & high MEMD & low entropy & high entropy \\
\midrule
$128\times 128$, $15\%$ patches & 46.7 & 50.5 & 46.2 & 52.7 \\
$128\times 128$, $30\%$ patches & 43.9 & 51.6 & 39.6 & 52.7 \\
\midrule
$256\times 256$, $15\%$ patches & 27.2 & 26.3 & 25.1 & 32.0 \\
$256\times 256$, $30\%$ patches & 45.5 & 57.2 & 52.7 & \textbf{71.0} \\
\bottomrule
\end{tabular}
\end{center}
\end{table}

The lower accuracy for the low MEMD and low entropy datasets, compared with the
high MEMD and entropy datasets, suggests that it is not sufficient to select
more patches to reach a higher level of accuracy; it is also important to select
appropriate patches.

\begin{figure}[h]
\centering
    \subfigure[Malignant lesion]{\includegraphics[width=.4\textwidth]{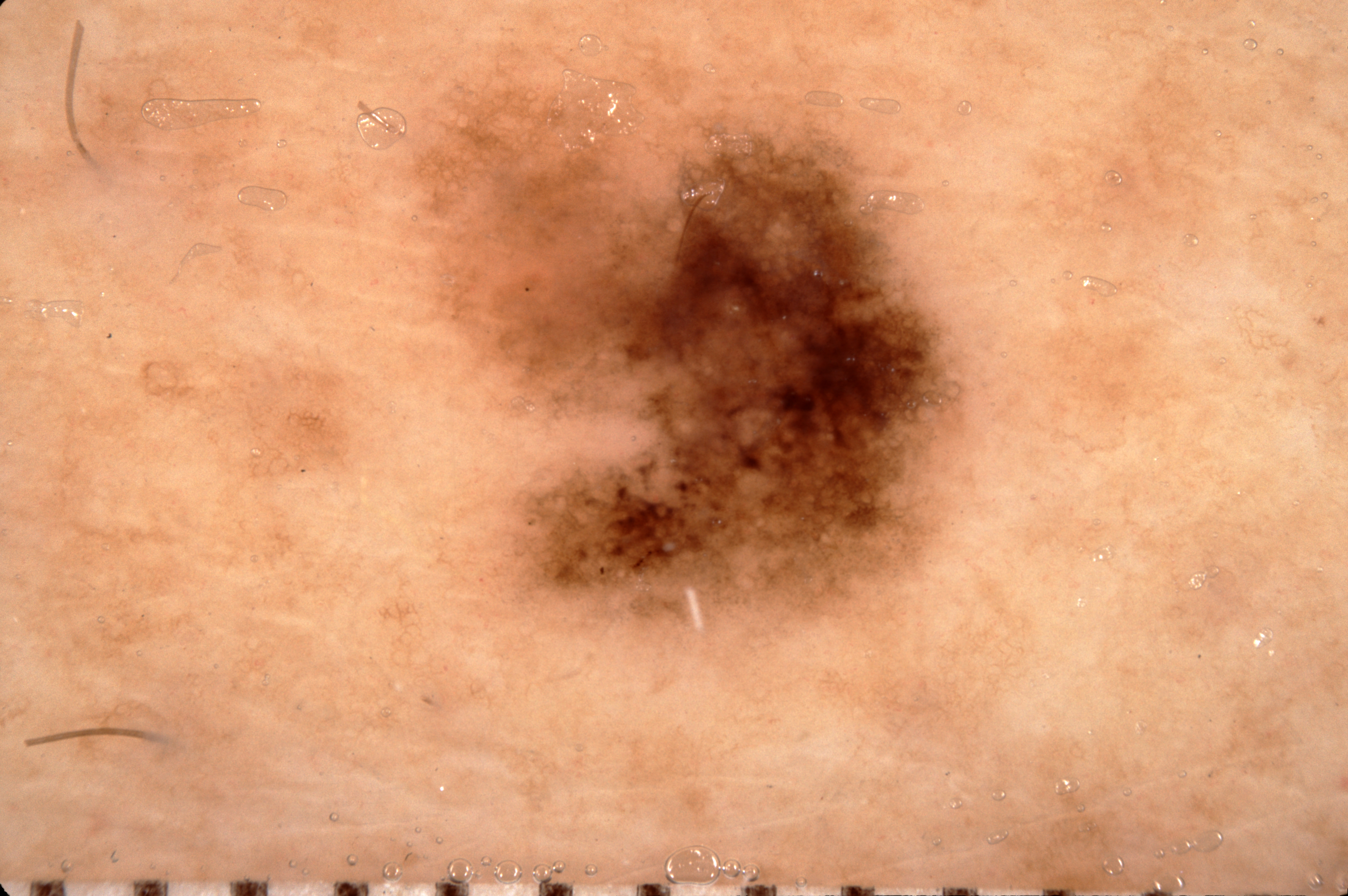}}
    \subfigure[Mask of the lesion]{\includegraphics[width=.4\textwidth]{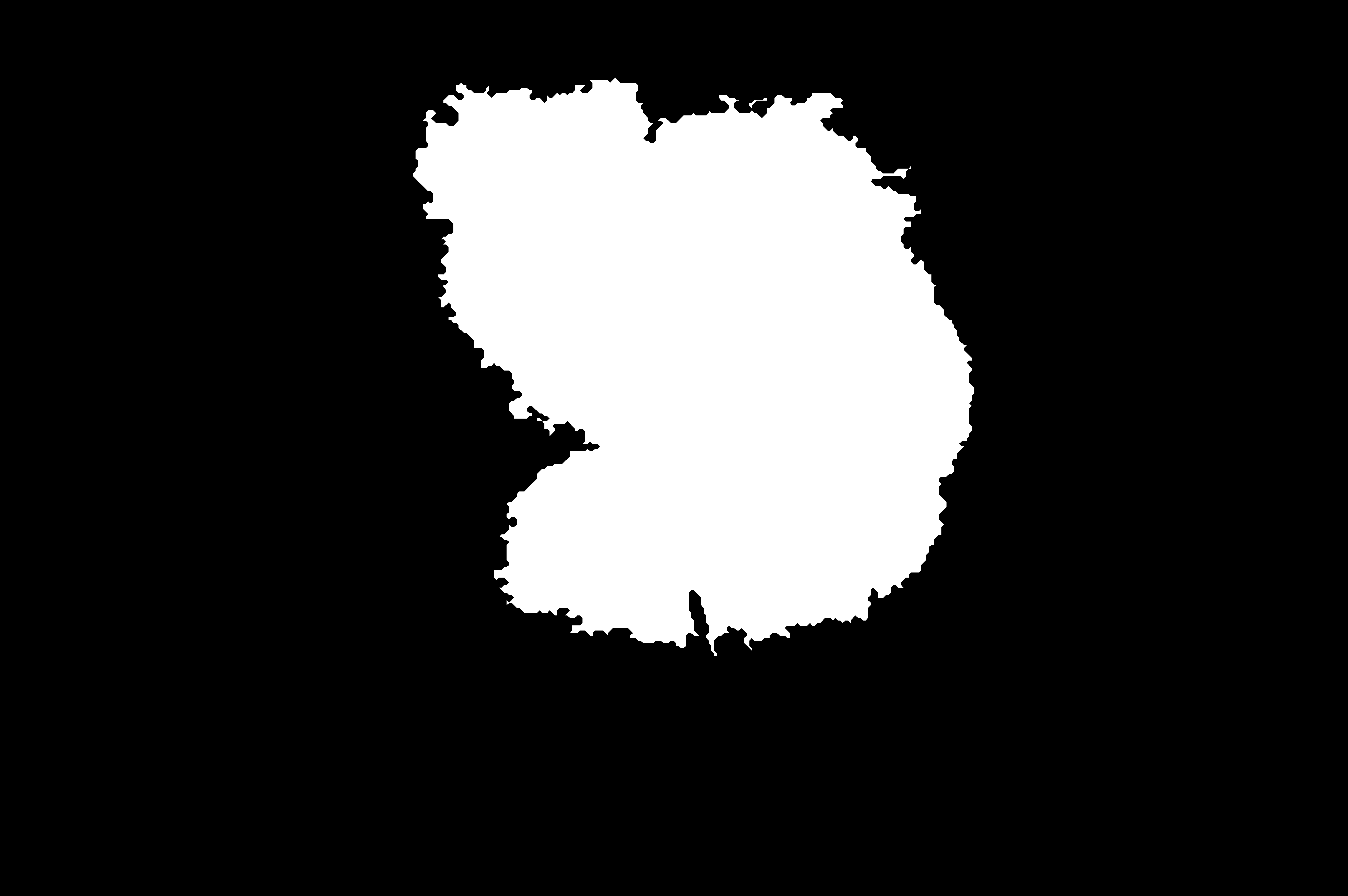}}
\caption{\centering A malignant image and its mask \label{fig5}}
\end{figure}

\begin{figure}[h]
\centering
    \subfigure[Patch with low MEMD score]{\includegraphics[width=.45\textwidth]{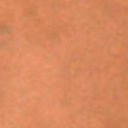}}
    \subfigure[Patch with high MEMD score]{\includegraphics[width=.45\textwidth]{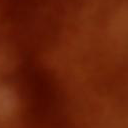}}
\caption{\centering Two different MEMD patches \label{fig6}}
\end{figure}

\begin{figure}[h]
\centering
    \subfigure[Patch with low entropy score]{\includegraphics[width=.45\textwidth]{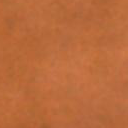}}
    \subfigure[Patch with high entropy score]{\includegraphics[width=.45\textwidth]{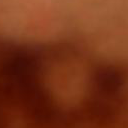}}
\caption{\centering Two different entropy patches \label{fig7}}
\end{figure}

Figure~\ref{fig6} and Figure~\ref{fig7} illustrate the role of MEMD and entropy in patch
selection. The patch on the left of Figure~\ref{fig6} is one of the patches with
the lowest MEMD score for the image, while the patch on the right has one of the
highest scores. Due to the fact that the masks cannot perfectly capture the
lesion, there will always be some part of the skin that will be present in the
mask. Since the skin has more uniform texture than the lesion, it is likely that
patches of skin will have the lowest score. Similarly, the patches with low entropy will have
more uniform features, while the patches with higher entropy will have more salient ones, as in Figure~\ref{fig7}.

The datasets extracted using the entropy converge faster than the datasets
extracted with the MEMD criterion. We hypothesize that a likely explanation is
that patches extracted with the entropy share similar distributions of pixels,
albeit sometimes shifted. The entropy quantifies the distribution of pixel
intensity: the higher the entropy, the closer the pixel distribution will be to
the uniform distribution.  Thus, the patches from entropy extracted datasets are
similar across the images, and this similarity is learnable by the network. On
the other hand, datasets extracted using the MEMD criterion do not provide any
quantifiable information about the pixel distribution. Their score is only
indicative of how representative the patch is with respect to the image. The
network might thus be confronted with a wider variety of patches which lead to a
longer training time.

\section{Conclusion}

We examined the role of entropy and the MEMD criteria on both CNN training time and classification efficiency for 
patch-based melanoma detection. The preprocessing is longer with
the MEMD criterion because we have to compare patches two by two, whereas
entropy requires a single computation per patch. We found that higher entropy
leads to faster convergence than lower entropy; similarly, a higher MEMD score,
which indicates that the patch does not resemble other patches from the same
image, also leads to faster convergence. In terms of accuracy, the models
trained on the higher entropy dataset or the higher MEMD are more performant
than the models trained on the lower entropy or lower MEMD datasets. We also
found that creating patch datasets using an absolute measure of information,
such as entropy, makes the network train faster than when the datasets were
created using a similarity measure. We also observed that patch size plays a
significant role in the classifier accuracy, with small patches leading to poor
results, regardless of the percentage of patches used.

\bibliographystyle{splncs04}
\bibliography{refs}

\end{document}